%% file: main.tex
\documentclass[10pt,twocolumn,letterpaper]{article}

\usepackage{cvpr}      

\input{preamble}

\definecolor{cvprblue}{rgb}{0.21,0.49,0.74}
\usepackage[pagebackref,breaklinks,colorlinks,allcolors=cvprblue]{hyperref}
\usepackage{booktabs}
\usepackage{multirow}
\usepackage{amsmath}
\usepackage{array}
\usepackage{graphicx} 
\usepackage{tabularx}
\usepackage{booktabs}
\usepackage{multirow}
\usepackage{cuted}
\usepackage{float}
\usepackage{fontawesome5}
\faStyle{solid|regular}
\usepackage[none]{hyphenat}
\newcolumntype{C}{>{\centering\arraybackslash}X}

\def\paperID{XXXX} 
\def\confName{CVPR}
\def\confYear{2026}

\title{Adapting Foundation Models for Annotation-Efficient Adnexal Mass Segmentation in Cine Images}

\author{
Francesca Fati$^{1,2}$,
Alberto Rota$^{2}$,
Adriana V. Gregory$^{1}$,
Anna Catozzo$^{1}$,\\
Maria C. Giuliano$^{1}$,
Mrinal Dhar$^{1}$,
Luigi De Vitis$^{1}$, 
Annie T. Packard$^{1}$, 
Francesco Multinu$^{3}$, \\
Elena De Momi$^{2,3}$, 
Carrie L. Langstraat$^{1},$ 
Timothy L. Kline$^{1,\text{\footnotesize{\faEnvelope}}}$ \\[6pt]
$^{1}$ Mayo Clinic,
$^{2}$ Politecnico di Milano,
$^{3}$ Istituto Europeo di Oncologia\\[6pt]
$^{\text{\faEnvelope}}$ 
\texttt{\small{Kline.Timothy@mayo.edu}}
}
\begin{document}

\maketitle
\begin{strip}
    \centering
    \vspace{-30pt}
    \includegraphics[width=\linewidth]{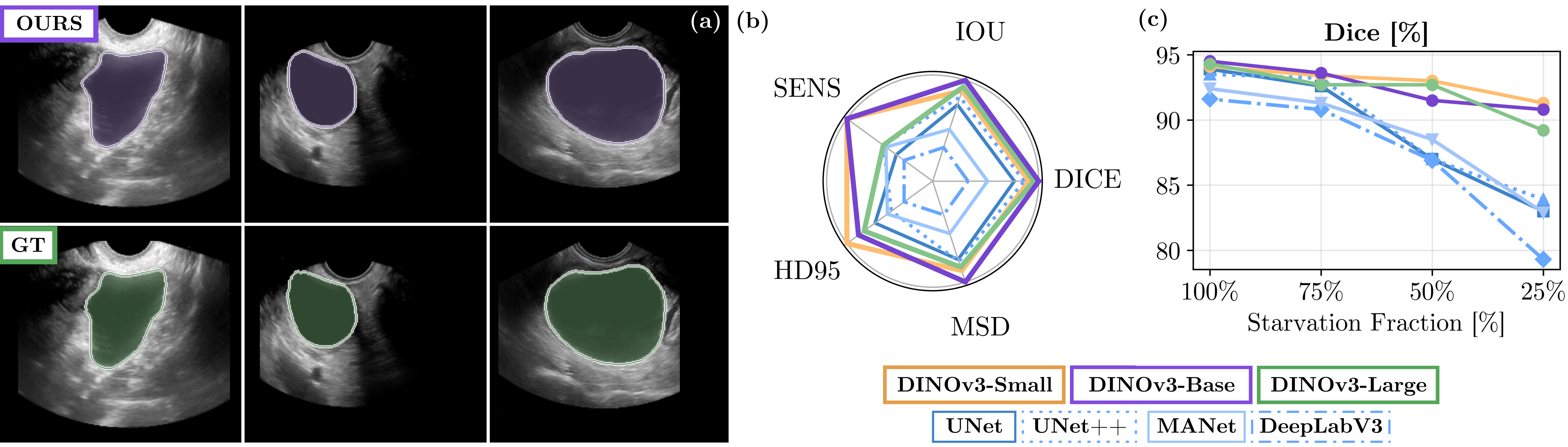}
    \captionof{figure}{\textbf{Adnexal mass segmentation from cine images}: \textit{(a)} Qualitative results demonstrating high boundary fidelity between the predicted masks and the ground truth. \textit{(b)} Performance comparison showing that DINOv3-based architectures outperform convolutional state-of-the-art baselines. \textit{(c)} Improved performance retention under data-starved training regimes.}
    \label{fig:header}
\end{strip}

\input{sec/0_abstract}    
\input{sec/1_intro}
\input{sec/2_related_works}

\input{sec/3_methodology}
\input{sec/4_experiments}
\input{sec/5_conclusion}

{
    \small
    \bibliographystyle{ieeenat_fullname}
    \bibliography{main}
}


\end{document}

%% file: preamble.tex
\usepackage{tabularx,array}
\usepackage{amsmath} 

\definecolor{ggreen}{HTML}{3EA944}
\definecolor{oorange}{HTML}{DF8820}
\definecolor{bblue}{HTML}{2971BC}
\definecolor{ppurple}{HTML}{7642CE}
\newcolumntype{Y}{>{\centering\arraybackslash}X}



%% file: sec/0_abstract.tex
\begin{abstract}
Adnexal mass evaluation via ultrasound is a challenging clinical task, often hindered by subjective interpretation and significant inter-observer variability. While automated segmentation is a foundational step for quantitative risk assessment, traditional fully supervised convolutional architectures frequently require large amounts of pixel-level annotations and struggle with domain shifts common in medical imaging. 
In this work, we propose a label-efficient segmentation framework that leverages the robust semantic priors of a pretrained DINOv3 foundational vision transformer backbone. By integrating this backbone with a Dense Prediction Transformer (DPT)-style decoder, our model hierarchically reassembles multi-scale features to combine global semantic representations with fine-grained spatial details. 
Evaluated on a clinical dataset of 7,777 annotated frames from 112 patients, our method achieves state-of-the-art performance compared to established fully supervised baselines, including U-Net, U-Net++, DeepLabV3, and MAnet. Specifically, we obtain a Dice score of 0.945 and improved boundary adherence, reducing the 95th-percentile Hausdorff Distance by 11.4\% relative to the strongest convolutional baseline. Furthermore, we conduct an extensive efficiency analysis demonstrating that our DINOv3-based approach retains significantly higher performance under data starvation regimes, maintaining strong results even when trained on only 25\% of the data. 
These results suggest that leveraging large-scale self-supervised foundations provides a promising and data-efficient solution for medical image segmentation in data-constrained clinical environments. 
\newline Project Repository:
\newline
\href{https://github.com/FrancescaFati/MESA}{\texttt{github.com/FrancescaFati/MESA}}
\end{abstract}

%% file: sec/1_intro.tex
\section{Introduction}
Adnexal masses are a common finding in gynecological practice, where the central clinical objective is to determine whether a lesion is benign or malignant \cite{liu2011management}. While most masses are benign, accurate characterization is critical, as delayed identification of malignancy is associated with poor outcomes \cite{rauh2011ovarian,badgwell2007early}. In routine workflows, this assessment relies on pelvic ultrasound (US), a widely available and noninvasive imaging modality \cite{benacerraf2015consider}. However, diagnostic performance depends strongly on acquisition quality and operator expertise, leading to substantial variability in interpretation \cite{koutras2023advantages}. Moreover, ultrasound images are inherently challenging due to speckle noise, acoustic shadowing, and low contrast, which complicate both manual and automated analysis \cite{baad2017clinical, whitney2024ai}.

Accurate segmentation is a key component of quantitative adnexal mass assessment. Reliable lesion delineation enables reproducible measurement of morphology, supports radiomics feature extraction, and facilitates downstream tasks such as malignancy risk prediction \cite{cai2018accurate,sahu2023comprehensive,barcroft2024machine}. However, manual annotation is time-consuming and subject to inter- and intra-observer variability \cite{timmerman1999subjective}, motivating the development of automated, annotation-efficient approaches.

In this work, we propose a segmentation framework for adnexal masses in ultrasound cine images that leverages a pretrained vision foundation model combined with a dense prediction head trained on domain-specific data. Our approach achieves state-of-the-art or competitive performance compared with established methods. To characterize data efficiency, we analyze performance under architectural variations and data-starvation regimes, providing a systematic evaluation of model capacity and training efficiency in this clinical setting.

This work makes three contributions, which we headline in Fig.~\ref{fig:header}.  First, we adapt a frozen DINOv3 backbone with a DPT-style decoder for adnexal mass segmentation in ultrasound cine images. Second, we benchmark this framework against established CNN baselines using patient-level splits and both region- and boundary-based metrics. Third, we provide a focused analysis of backbone scale, input resolution, and training-data starvation.

%% file: sec/2_related_works.tex
\section{Related Work}
Accurate characterization of adnexal masses is clinically consequential: malignant lesions should be referred promptly for appropriate oncologic management, whereas benign or asymptomatic masses should ideally be spared unnecessary intervention. In routine practice, however, diagnostic performance depends strongly on operator experience and subjective interpretation of ultrasound findings, resulting in substantial inter- and intra-observer variability \cite{whitney2024ai,barcroft2024machine}. These limitations have motivated computer-aided diagnosis pipelines for adnexal ultrasound, particularly in settings where expert sonographers are scarce and clinical volume is high \cite{li2022deep,barcroft2024machine}.

A consistent theme across this literature is that    \emph{segmentation} is a prerequisite for downstream quantitative modeling. Reliable delineation enables reproducible measurement of lesion geometry, supports radiomic feature extraction, and reduces reliance on time-consuming manual contouring. 
Whitney et al.\ proposed an AI-based pipeline for ultrasound images in which adnexal mass segmentation is the first step, followed by component-level analysis to support radiomics and workflow efficiency \cite{whitney2024ai}. Their study underscores the practical challenges of segmentation in ultrasound images, specifically speckle noise, acoustic shadowing, and heterogeneous lesion appearance, all of which can obscure boundaries and degrade generalization. Liu et al.\ integrated automatic segmentation with ultrasound radiomics and machine-learning classifiers to stratify lesions into low-risk versus intermediate/high-risk groups, illustrating how segmentation quality can directly influence downstream clinical decisions \cite{liu2025automatic}. Similarly, a recent study proposed a deep learning framework that leverages nnU-Net for automated segmentation of adnexal masses, predicting fluid and solid components from entire B-mode ultrasound images. The segmented outputs are integrated into a multimodal classification network to differentiate benign from malignant lesions \cite{dhar2026adnexal}. 
Barcroft et al. systematically evaluated an end-to-end segmentation–classification framework, comparing commonly used deep segmentation architectures for the segmentation component \cite{barcroft2024machine}. Specifically, they examined U-Net and widely used variants: (i) U-Net, a symmetric encoder--decoder with skip connections that preserve fine spatial detail; (ii) U-Net++, which introduces nested skip pathways and dense connectivity to improve multi-scale feature fusion; (iii) DeepLabV3, which employs \textit{à trous} spatial pyramid pooling  to aggregate multi-scale context (typically with a ResNet backbone); and (iv) MANet, an attention-augmented encoder--decoder that enhances feature representation via attention modules. Their results indicate that a U-Net-based model achieved the strongest segmentation performance in their setting \cite{barcroft2024machine}. 

Nevertheless, data scarcity and class imbalance remain persistent barriers; Lebbos et al.\ highlighted the difficulty of training reliable ultrasound segmenters under skewed pathology distributions and emphasized the need for strategies that mitigate limited annotations and under-represented classes \cite{lebbos2022adnexal}.
More broadly, adnexal mass ultrasound segmentation has predominantly relied on fully supervised convolutional encoder--decoder models, with U-Net and its variants serving as strong baselines \cite{ronneberger2015unet, wanderley2018end, marques2019segmentation}. However, purely supervised CNN pipelines often require substantial amounts of high-quality pixel-level annotations \cite{hoffman2016fcns}, careful augmentation, and extensive retuning to maintain performance across heterogeneous acquisition conditions and domain shifts, which are common in ultrasound \cite{lebbos2022adnexal}. The relationship between training set size and CNN performance is a critical bottleneck in medical imaging; as demonstrated by Gottlich \textit{et al.}, model accuracy is heavily contingent on data scale and follows modality-specific scaling laws \cite{gottlich2023effect}.

Transformer-based dense prediction has emerged as an alternative that can better exploit global context. DPT is a representative framework that aggregates multi-scale transformer representations into full-resolution dense outputs and has become a widely adopted template for pixel-wise prediction \cite{ranftl2021dpt}. In parallel, self-supervised pretraining has recently demonstrated the ability to learn transferable visual representations: DINO \cite{caron2021dino} showed that self-distilled vision transformer representations encode meaningful spatial structure even without labels, and DINOv3 further targets \emph{robust dense} features, making it particularly relevant for boundary-sensitive segmentation under limited annotation and strong appearance variability \cite{simeoni2025dinov3, krishnan2022self, chowdhury2021applying}. 

Extending these advancements to adnexal mass segmentation, we introduce a framework that leverages DINOv3 self-supervised representations via a DPT-style decoder. By coupling a pretrained foundation backbone with a task-specific dense prediction head, we achieve robust performance that scales effectively under data scarcity.

%% file: sec/3_methodology.tex
\section{Methodology}
\label{sec:methodology}
We propose a deep learning framework for the automated segmentation of adnexal masses in ultrasound images. Our architecture integrates the DINOv3 \cite{simeoni2025dinov3} foundational backbone with a high-performance dense prediction head based on the DPT \cite{ranftl2021dpt} architecture. The model is trained using a fully supervised paradigm on a dataset of manually annotated cine frames containing expert-derived segmentation masks. 

\subsection{Model Architecture}
\begin{figure*}
    \centering
    \includegraphics[width=1\linewidth]{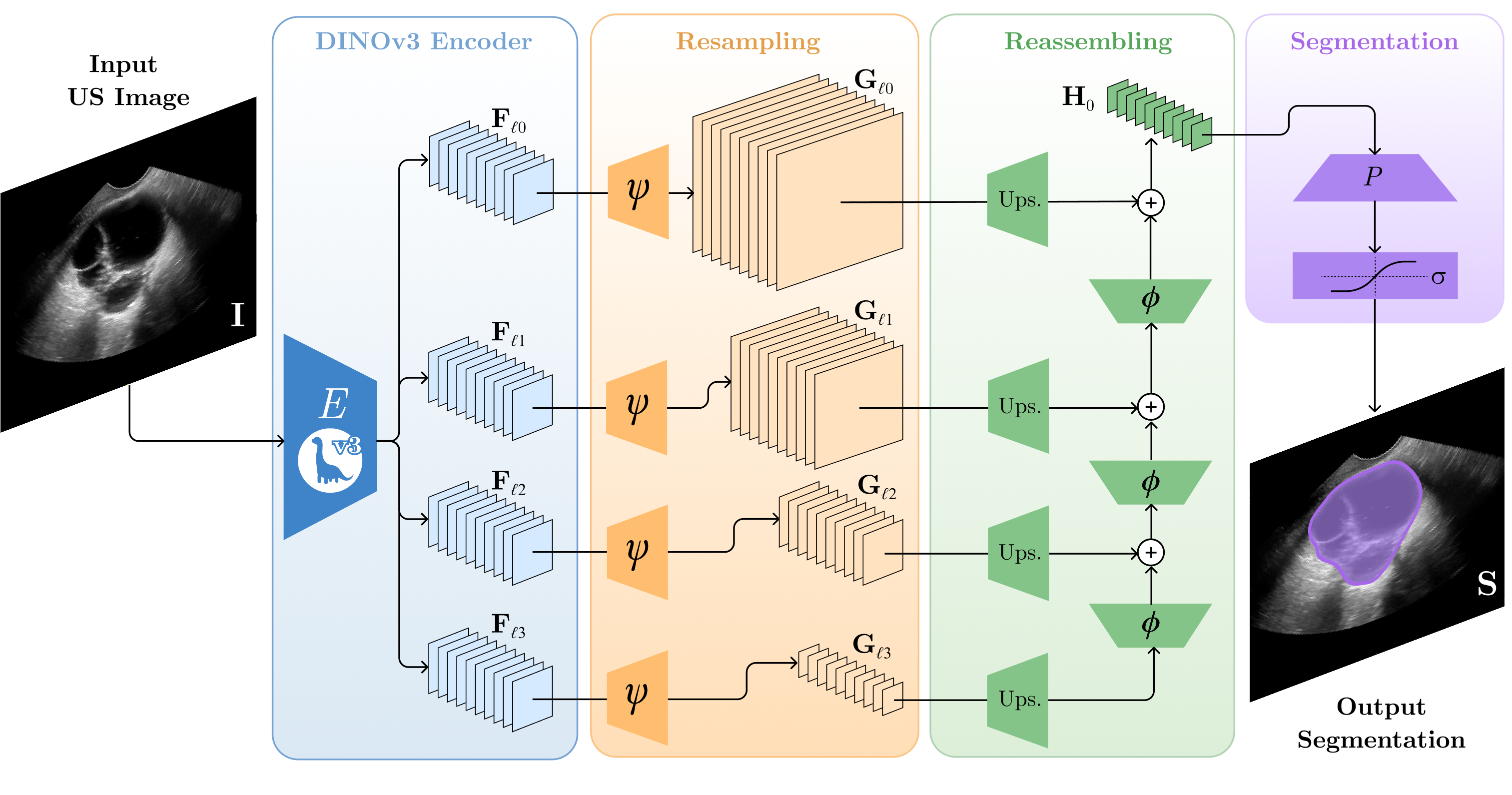}
    \caption{
    \textbf{Model Architecture Overview}: The input image $\mathbf{I}$ is processed by the \textcolor{bblue}{\textbf{DINOv3 Encoder} $E$} which extracts a set of hierarchical feature maps from which we retain only the subset $\{\mathbf{F}_{\ell_0}, \mathbf{F}_{\ell_1}, \mathbf{F}_{\ell_2}, \mathbf{F}_{\ell_3}\}$. The feature maps are then passed to a learned \textcolor{oorange}{\textbf{resampling operator $\psi$}} that resizes them to a higher or lower resolution depending on their rank in the hierarchy, $\{\mathbf{G}_{\ell_0}, \mathbf{G}_{\ell_1}, \mathbf{G}_{\ell_2}, \mathbf{G}_{\ell_3}\}$. Later, the \textcolor{ggreen}{\textbf{Upsample}} and \textcolor{ggreen}{\textbf{reassemble modules} $\phi$} progressively fuse the resampled maps into a final representation that is processed by the \textcolor{ppurple}{\textbf{Segmentation head} $H$,} to obtain the final segmentation mask $\mathbf{S}$.
    }
    \label{fig:modelarch}
\end{figure*}

Given a grayscale cine frame $\mathbf{I} \in \mathbb{R}^{H \times W}$, our model $M$ is designed to predict a binary segmentation mask $\mathbf{S} \in \{0,1\}^{H \times W}$ that delineates the adnexal mass region. The model $M$ consists of two primary components: a pretrained feature extractor $E$  and a dense prediction decoder $D$ that hierarchically reassembles and upscales the extracted features into a spatial binary mask. The model architecture is represented in Fig. \ref{fig:modelarch}.
Formally:
\begin{equation}
    \mathbf{S} = M(\mathbf{I}) = D\left(E\left(\mathbf{I}\right)\right)
\end{equation}

The encoder $E$ is the vision foundation model DINOv3, pretrained on a massive corpus of natural RGB images to learn robust semantic priors. We leverage its documented cross-domain adaptability \cite{li2025meddinov3adaptvisionfoundation, simeoni2025dinov3} to extract high-level semantics from medical cine frames. Since the encoder $E$ requires three-channel inputs, we convert each grayscale US image into a pseudo-RGB format via channel replication.

$E$ maps the input image $\mathbf{I}$ to a set of $L$ hierarchically organized feature maps, denoted as:
\begin{equation}
\mathcal{F} = \{\mathbf{F}_0, \mathbf{F}_1, \mathbf{F}_2, \dots, \mathbf{F}_L\} = E(\mathbf{I})
\end{equation}
where each $\mathbf{F}_\ell \in \mathbb{R}^{E_d \times \frac{H}{P_s} \times \frac{W}{P_s}}$. Here, $E_d$ represents the embedding dimension and $P_s$ denotes the patch size. Consequently, the feature maps in $\mathcal{F}$ have a lower spatial resolution than the original input. In addition to spatial features, DINOv3 extracts one classification token and four register tokens within the $\mathbb{R}^{E_d}$ vector space. While these tokens encode global context, they are discarded for this pixel-level localization task to focus exclusively on spatial representations.

The hierarchical feature maps from $\mathcal{F}$ are fused using the decoder $D$ as an adaptation of the Dense Prediction Transformer architecture, implemented in a RefineNet \cite{lin2016refinenetmultipathrefinementnetworks} style. Our decoder processes a specific subset $\{\mathbf{F}_{\ell_0}, \mathbf{F}_{\ell_1}, \mathbf{F}_{\ell_2}, \mathbf{F}_{\ell_3}\}$ of four feature maps sampled at different depths of the hierarchy. This design choice is predicated on the fact that different layers in $\mathcal{F}$ capture varying levels of information: shallower layers preserve granular details such as intensity and texture, while deeper layers encode complex structural and semantic context. By aggregating features across multiple hierarchies rather than relying solely on the final hidden state, the model ensures the preservation of both local boundary precision and global semantic consistency.

The decoding process follows three conceptual stages: resampling, reassembly, and prediction.

\paragraph{Feature Resampling.}
The selected feature maps from the frozen encoder are first projected into a common task-specific embedding space. During this stage, we apply learned resampling operators $\psi(\cdot )$ to adjust the spatial resolution of each map, creating a feature pyramid 

\begin{equation}
    \{\mathbf{G}_{\ell_k}\}_{k=1}^4 = \psi \left( \{\mathbf{F}_{\ell_k}\}_{k=1}^4 \right)
\end{equation} 

that represents the image at multiple effective scales.

\paragraph{Reassembling and Fusion.}
The feature pyramid is integrated from the coarsest to the finest resolution using a sequence of residual fusion blocks. Each block $\phi(\cdot)$ refines the incoming feature map to enhance spatial details. Starting from the lowest resolution, the decoder iteratively upsamples the fused representation and merges it with the next refined map in the pyramid:
\begin{equation}
    \mathbf{H}_{\ell_{k+1}} = \mathrm{Upsample}(\mathbf{G}_k) + \phi \left( \mathbf{H}_{\ell_{k+1}} \right)
\end{equation}
where $\mathbf{H}_k$ is the integrated feature state at scale $k$. This merger effectively guides deep-layer semantic information using spatial details from early layers to maintain structural integrity.

\paragraph{Prediction.}
The final integrated feature map $\mathbf{H}_0$, which now contains a rich combination of multi-scale information, is passed through a lightweight convolutional prediction head $P$. This head performs a final spatial upsampling to match the original image dimensions and applies a sigmoid activation $\sigma(\cdot)$ to generate the binary probability mask $\mathbf{S}$, formally
\begin{equation}
    \mathbf{S} = \sigma \left( P(\mathbf{H}_0) \right)
\end{equation}
\begin{figure}[h]
    \centering
    \includegraphics[width=1\linewidth]{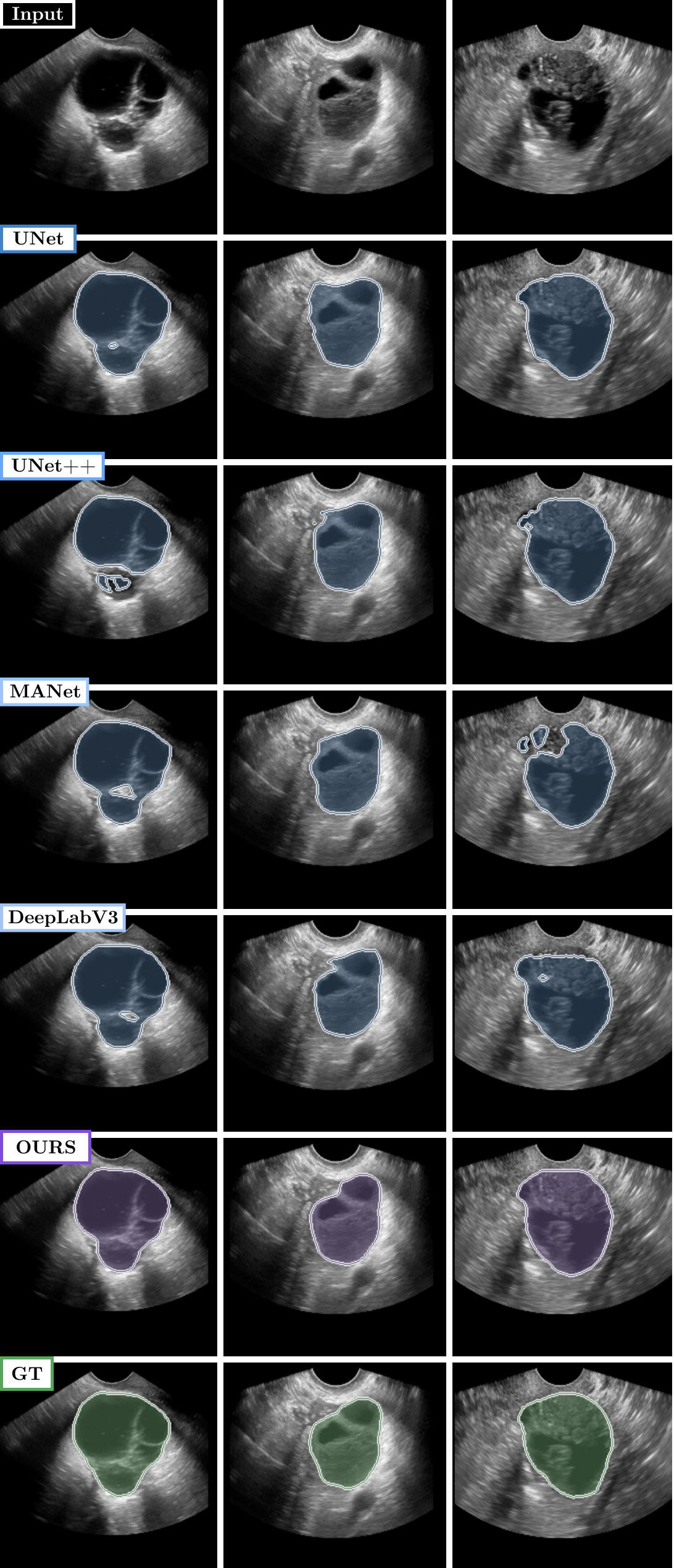} 
    \caption{\textbf{ Qualitative Segmentation Comparison}: Visual comparison of adnexal mass segmentation results across different architectures. Our method demonstrates superior boundary fidelity and structural consistency.} 
    \label{fig:qualitative}
\end{figure}
\subsection{Supervision}
We trained our model in a supervised fashion using a combination of a per-pixel loss formulated as a Binary Cross-Entropy and a region-overlap criterion formulated as a Dice loss to address class imbalance and boundary ambiguity common in US images. 
The total loss is therefore computed as the weighted average
\[
\mathcal{L} = \lambda_{\text{BCE}}\,\mathcal{L}_{\text{BCE}}(\hat{Y},Y)
+ \lambda_{\text{Dice}}\,\mathcal{L}_{\text{Dice}}(\hat{Y},Y).
\]
where the Dice loss is expressed as:
\[
\mathcal{L}_{\text{Dice}} = 1 - \frac{2\sum_i \hat{y}_i y_i + \epsilon}{\sum_i \hat{y}_i + \sum_i y_i + \epsilon},
\]
with a small $\epsilon$ for numerical stability. The DINOv3 backbone is kept frozen in its pre-trained state at training time.

%% file: sec/4_experiments.tex
\section{Experiments}
\begin{table*}
\centering
\small
\setlength{\tabcolsep}{2pt} 
\caption{\textbf{Quantitative comparison of segmentation performance} of adnexal mass segmentation performance across the related work at resolutions $224\times 224$ and $512\times 512$. For '$\uparrow$', higher is better, for '$\downarrow$' lower is better: bold typeface indicates the best performing method.}
\begin{tabularx}{\linewidth}{@{}l|CC|CC|CC|CC|CC@{}}
\toprule
\multirow{2}{*}{Method} & \multicolumn{2}{c|}{Dice $\uparrow$} & \multicolumn{2}{c|}{IoU $\uparrow$} & \multicolumn{2}{c|}{Sens. $\uparrow$} & \multicolumn{2}{c|}{HD95 $\downarrow$} & \multicolumn{2}{c}{MSD $\downarrow$} \\
 & 224 & 512 & 224 & 512 & 224 & 512 & 224 & 512 & 224 & 512 \\
\midrule
U-Net     & 0.939 & 0.880 & 0.884 & 0.785 & 0.942 & 0.833 & 11.13  & 65.886  & 2.664 & 20.405 \\
U-Net++   & 0.935 & 0.875 & 0.878 & 0.778 & 0.937 & 0.823 & 10     & 65     & 2.703 & 20.143 \\
DeepLabV3 & 0.916 & 0.932 & 0.845 & 0.872 & 0.934 & 0.949 & 12     & 18.868     & 3.526 & 5.842 \\
MAnet     & 0.924 & 0.932 & 0.859 & 0.873 & 0.941 & 0.953 & 10.851 & 21.201 & 3.18  & 6.378  \\
\midrule
\textbf{OURS} & \textbf{0.945} & \textbf{0.937} & \textbf{0.897} & \textbf{0.882} & \textbf{0.956} & \textbf{0.956} & \textbf{8.861} & \textbf{18.455} & \textbf{2.288} & \textbf{5.323} \\
\bottomrule
\end{tabularx}
\label{tab:quantitivecomparison}
\end{table*}
\label{sec:experiments}
We conduct all experiments on a NVIDIA A100 GPU with 80GB RAM. In our best experiments, the image dimensions are $H=224$ and $W=224$, the embedding dimension of DINOv3-Base is $E_d=786$ and it extracts feature maps at $L=12$ hierarchical layers with patch size $P_s=14$. We extract $\{\mathbf{F}_3, \mathbf{F}_6, \mathbf{F}_9, \mathbf{F}_{12}\}$ to be reassembled in the decoder.
  We train the model for 100 epochs with the AdamW optimizer with using learning rate $\eta{=}1.0\cdot 10^{-4}$ annealed with a cosine schedule, a weight decay $w_d{=}1.0\cdot 10^{-3}$ and a batch size of $B=128$. The loss function weights are $\lambda_{\text{BCE}}=0.3$ and $\lambda_{\text{Dice}}=0.7$. One training run reaches convergence after approximately 8 GPU hours.

\subsection{Dataset}
 Our study is based on a clinical dataset of adnexal mass ultrasound cine clips acquired using both transvaginal and transabdominal approaches. Expert clinicians provided manual pixel-level annotations for all lesions in a cohort of 112 patients, comprising 9,923 frames in total (7,777 with annotated lesions and 2,146 background frames). To prevent data leakage, a strict patient-level partitioning strategy was applied, ensuring that all frames from a given patient were assigned exclusively to a single dataset split. The data were stratified by benign and malignant cases and divided into training, validation, and test sets in a 70/15/15 ratio, with no patient overlap between subsets.
The dataset reflects significant morphological heterogeneity, encompassing both benign and malignant pathologies. Input frames were resized to either 224×224 or 512×512 pixels and underwent Z-score normalization. 

\subsection{Comparison with the Related Work}

\begin{figure*}[h]
    \centering
    \includegraphics[width=\linewidth]{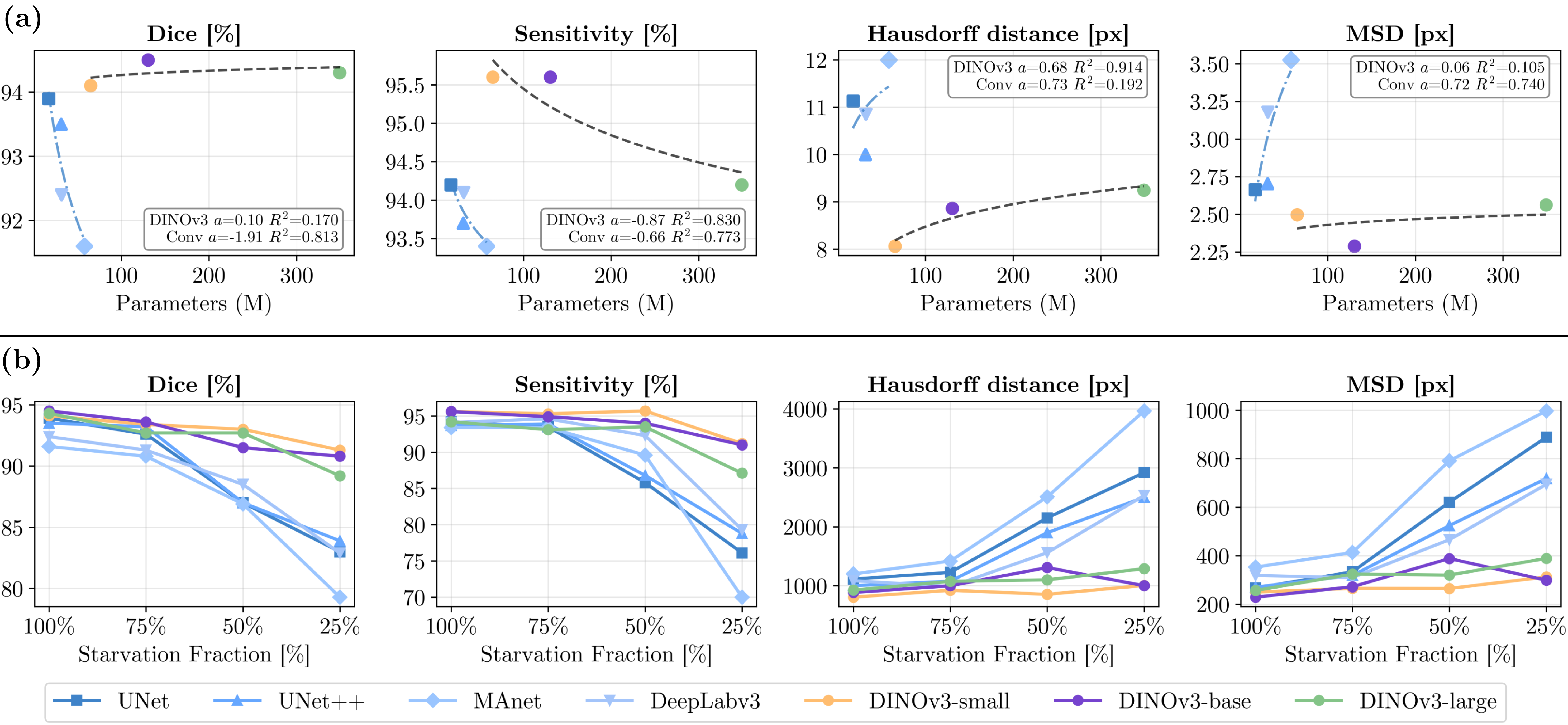}
    \caption{\textbf{Efficiency Analysis Results}: \textit{a)} \textit{Pareto efficiency analysis} comparing segmentation performance against model capacity. We fit a logarithmic curve to the methods based on DINOv3 and to the convolutional state of the art separately and we report the log slope $a$ and the fit quality $R^2$. 
    \textit{b)} \textit{Data Starvation Curves} report the performance losses at progressively larger starvation fractions.
    }
    \label{fig:efficiency}
\end{figure*}

We evaluate our framework against established medical segmentation architectures, including U-Net, U-Net++, DeepLabV3, and MANet. To ensure a rigorous and fair comparison, we utilize domain-specific implementations from the open-source adnexal mass analysis suite\footnote{\url{https://github.com/Cancer-Imaging-Group/adnexal-mass-classifier}}, providing a standardized benchmark for this clinical task. All models are trained using identical preprocessing pipelines and training protocols. Each network outputs a binary probability map, and performance is assessed using a per-image evaluation protocol. 

Segmentation accuracy is quantified using both overlap-based and distance-based metrics.  Per-pixel classification performance is evaluated through Sensitivity (Sens.).
Overlap accuracy is quantified using the Dice similarity coefficient (Dice) and Intersection over Union (IoU). 
To  assess boundary accuracy and spatial agreement between predicted and reference segmentations, we compute the Hausdorff distance at the 95-\textit{th} percentile and the Mean Surface Distance (MSD) as distance-based metrics, both quantified in pixels.
HD95 measures the 95th percentile of the bidirectional distances between segmentation boundaries, reducing sensitivity to outlier pixels and providing a robust measure of the "worst-case" error \cite{lebbos2022adnexal}. MSD quantifies the average symmetric distance between corresponding surface points of the predicted and ground-truth masks, capturing the overall boundary deviation and providing a stable measure of contour agreement \cite{lei2019ultrasound}.

The quantitative results of this comparative analysis are reported in Table~\ref{tab:quantitivecomparison}, while a qualitative comparison is displayed in Figure~\ref{fig:qualitative}. At lower resolution, the proposed method demonstrates consistent improvements over the evaluated baselines across the majority of metrics. In terms of spatial overlap, our approach achieves the highest Dice similarity coefficient and IoU, yielding relative performance increases of approximately +0.6\% and +1.3\% over U-Net, which served as the most competitive baseline. Sensitivity also improves by nearly +1.4\%, indicating better detection of the target region. The distance-based metrics further highlight improved boundary adherence. Our model reduces the 95\textit{th}-percentile Hausdorff Distance by 11.4\% compared to the next best method (U-Net++), reflecting fewer severe boundary errors. Similarly, the Mean Surface Distance is reduced by 14.1\% relative to U-Net, suggesting a closer overall alignment of the predicted contours with the ground truth. Similar comparative performance is exhibited at a higher resolution.

\subsection{Efficiency Analysis}

Data scarcity and a lack of dataset heterogeneity are critical bottlenecks for data-driven approaches in clinical applications. Vision foundation models pre-trained on massive "in-the-wild" datasets help bridge this gap. However, these models are available in multiple distillations, leaving it unclear which one has the optimal capacity for specific clinical tasks. A smaller model may fail to capture necessary semantic features and consequently require significantly more data or gradient steps to converge. Conversely, a larger model might offer superior few-shot discriminability and require only a lightweight prediction head, but introduces the risk of excessive computational overhead during fine-tuning.

To resolve these trade-offs, we investigate how segmentation performance varies in relation to architectural complexity, input spatial resolution, and the volume of available training data. We analyze these dynamics through the following test cases.

\subsubsection{Backbone Scale} 
We progressively increase the capacity of the DINOv3 backbone from \textit{Small} ($E=384$) to \textit{Base} ($E=768$) and \textit{Large} ($E=1024$) to assess the performance gains provided by deeper representations. 

Fig~\ref{fig:efficiency}.a reports the results of our analysis, which reveals a rapid capacity saturation effect. By analyzing the logarithmic scaling slope of performance metrics against parameter count, we observe that representational scaling does not translate monotonically into clinical segmentation accuracy. Transitioning from the \textit{Small} ($\sim$65M parameters) to the \textit{Base} architecture ($\sim$130M parameters) yields only marginal improvements in volumetric overlap, with Dice increasing from 0.941 to 0.945 and MSD improving from 2.497 to 2.288. However, the log slope flattens and ultimately becomes negative when scaling further to the \textit{Large} configuration ($\sim$349M parameters), which exhibits a degradation across both overlap and boundary distances.
The \textit{Large} backbone distillation might be prone to overfitting the fine-tuning dataset.

\subsubsection{Data Starvation} We train our DINOv3-based segmentation model on progressively smaller, randomly sampled subsets of the training dataset, systematically undersampling the training pool at \textit{starvation fractions}: 100\% (95 patients), 75\% (71 patients), 50\% (47 patients), and 25\% (23 patients). The test set remains unchanged to the same 17 patients.

We quantify performance in this experiment with the Retention Index $RI_m$, for each of the $m$ performance metrics. We compute it as the ratio 
\begin{equation}
    RI_m = \frac{m_{100\%}}{m_{25\%}}
\end{equation}
(or its inverse for metrics where higher performance is better like HD95 and MSD) to quantify how well each method retains its performance when starved to 25\% of its learning data. The quantitative results are collected in Tab.~\ref{tab:starvation} and the full trends at all starvation percentages as displayed in Fig.~\ref{fig:efficiency}.b. 
\begin{table}[t]
\centering
\small
\setlength{\tabcolsep}{6pt}
\caption{Retention Indexes for all metrics under -75\% data starvation. Best performance is in bold typeface}
\begin{tabularx}{\linewidth}{@{}l|YYYYY} 
\toprule
  & \multicolumn{5}{c}{Retention Index $RI$} \\ 
Method & Dice & IoU & Sens & HD95 & MSD \\
\midrule
UNet & 0.884 & 0.803 & 0.808 &  0.381 & 0.300 \\
UNet++ & 0.897 & 0.823 & 0.841 &  0.400 & 0.376 \\
DeepLabv3 & 0.866 & 0.779 & 0.749 &  0.303 & 0.353 \\
MAnet & 0.897 & 0.824 & 0.843 &  0.429 & 0.458 \\
\midrule
DINOv3-small & \textbf{0.970} & \textbf{0.944} & \textbf{0.954} &  0.802 & \textbf{0.800} \\
DINOv3-base & 0.961 & 0.926 & 0.952 & \textbf{0.886} & 0.766 \\
DINOv3-large & 0.946 & 0.902 & 0.925 &  0.717 & 0.659 \\
\bottomrule
\end{tabularx}

\label{tab:starvation}
\end{table}
Results clearly suggest that architectures based on DINOv3 are capable of retaining strong performance across all metrics when starved up to -75\% of the training data, while the convolutional related works struggle to maintain even 90\% of the performance obtained learning from the whole dataset. Curiously, the \textit{small} distillation seems to be marginally more robust compared to the more parameter-heavy ones. 
\begin{figure}[t]
    \centering
    \includegraphics[width=\linewidth]{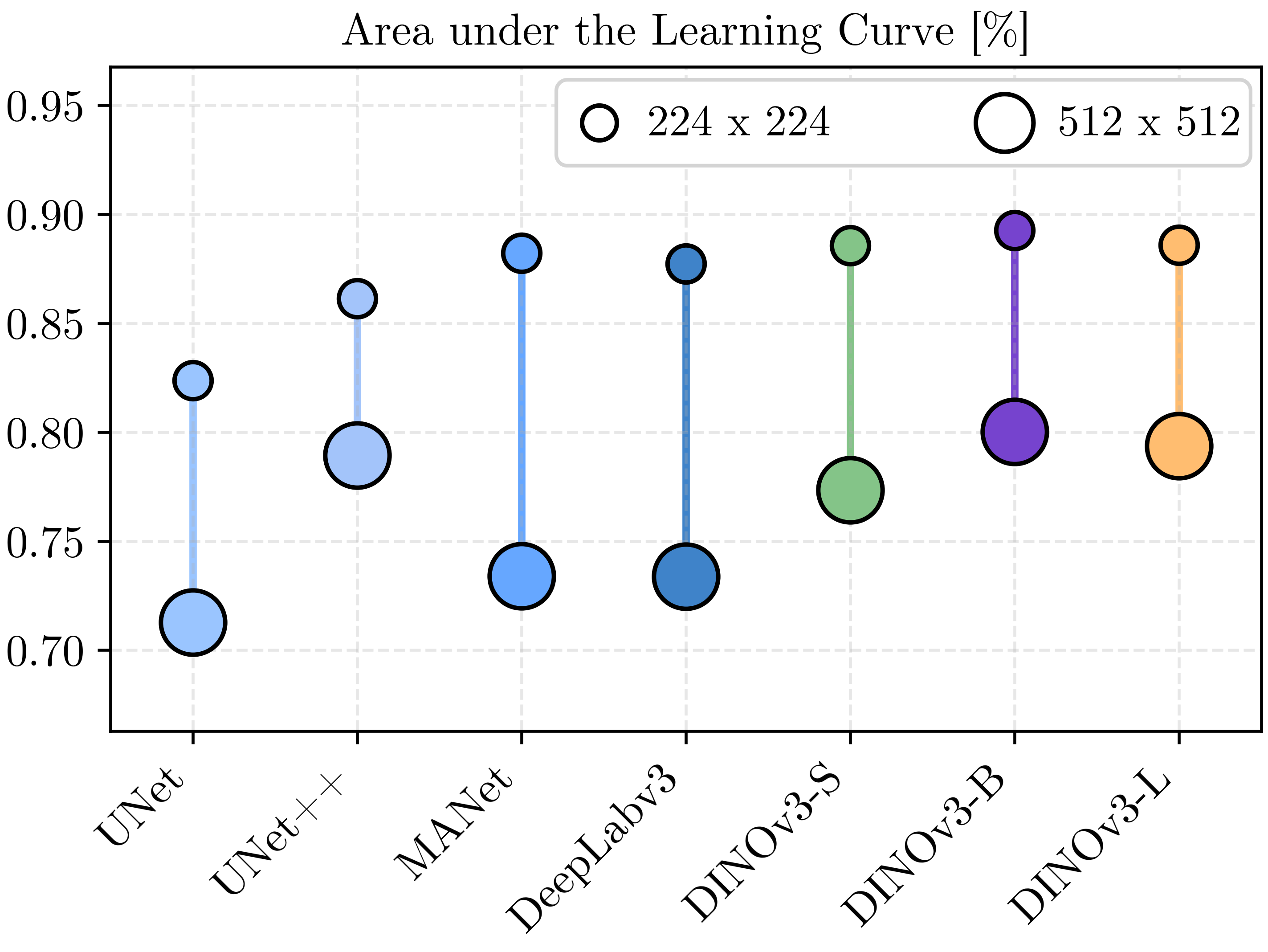}
    \caption{\textbf{Convergence efficiency across architectures.} Area under the Learning Curve (ALC) for different models and backbone sizes at two input resolutions. Each pair of markers shows the same architecture at different resolutions.}
    \label{fig:alc}
\end{figure}

\subsubsection{Resolution Effects on Learning Quality}
We scale the default DINOv3 preprocessing dimensions from $224\times224$ to the higher $512\times512$ to determine if increased spatial resolution provides a benefit in terms of learning speed. In principle, higher resolution should facilitate the delineation of fine-grained adnexal structures; however, our empirical results suggest a contrary phenomenon in the context of frozen foundation models.

As illustrated in Fig.~\ref{fig:alc},
increasing the input resolution to $512\times512$ yields a consistent and substantial degradation in the Area under the Learning Curve (ALC) across all backbone scales. For instance, the DINOv3-small variant experiences a relative drop of 10.4\% in ALC when moving to the higher resolution. This inverse relationship indicates that the larger input space complicates the optimization landscape.
The plot shows that the performance degradation is consistent regardless of model scale. This suggests the bottleneck is not the capacity of the model, but the representation alignment. The $512\times512$ resolution likely creates a flatter, more complex optimization landscape where the gradient signal from the frozen backbone is less informative at each iteration.

%% file: sec/5_conclusion.tex
\section{Discussions}
Our findings demonstrate that dense semantic priors derived from DINOv3 substantially improve the accuracy and in-domain robustness of adnexal mass segmentation in ultrasound imaging compared with convolutional state-of-the-art models. By coupling a frozen DINOv3 backbone with a DPT-style multi-scale decoder, the proposed architecture consistently outperforms established convolutional baselines across both overlap- and boundary-based metrics. Gains in Dice and IoU, together with significant reductions in HD95 and MSD, indicate not only improved volumetric agreement but also more anatomically consistent contour delineation.

The most interesting findings come from our efficiency analysis, which reveals two salient phenomena. First, we observe rapid capacity saturation when scaling backbone size. In contrast to the power-law scaling behavior commonly reported in natural image domains, the Large DINOv3 variant exhibits performance degradation, with signs of overfitting. This suggests that ultrasound segmentation operates in a regime where excessive representational capacity amplifies dataset-specific noise and acquisition artifacts rather than improving structural generalization. The Small and Base variants instead define a Pareto-optimal frontier, achieving a favorable balance between parameter efficiency and spatial accuracy. Second, DINOv3-based models demonstrate pronounced robustness under severe data starvation. When trained on only 25\% of the available data, these models retain substantially higher performance than fully supervised CNN baselines. This resilience underscores the practical value of self-supervised priors in annotation-constrained medical environments, where high-quality pixel-level labels are costly, time-intensive, and inherently variable. The observed retention behavior suggests that foundation-model representations act as a strong structural regularizer, constraining the hypothesis space and mitigating overfitting under limited supervision.

Collectively, these results indicate that the principal advantage of foundation models in the context of adnexal mass segmentation lies not merely in increased representational capacity, but in the imposition of a structured semantic prior that stabilizes learning under domain heterogeneity, speckle noise, and boundary ambiguity, hallmarks of ultrasound imaging.

Several limitations merit consideration. This evaluation is limited to internal validation on a single-institution cohort and should not be interpreted as evidence of cross-center generalization. Furthermore, cine frames were treated as independent samples, potentially underestimating temporal redundancy and inflating the effective sample size. Rigorous multi-center validation across scanners, acquisition protocols, and patient populations is necessary to fully characterize generalization. Most importantly, the downstream impact on malignancy risk stratification and clinical decision-making remains to be systematically evaluated.

\section{Conclusions}
We presented a label-efficient framework for adnexal mass segmentation that integrates a self-supervised DINOv3 foundation backbone with a Dense Prediction Transformer decoder. Quantitative evaluation demonstrates state-of-the-art performance relative to established fully supervised CNN architectures, with consistent gains across overlap-based and boundary-sensitive metrics. Notably, the observed reductions in Hausdorff distance and mean surface distance underscore improved contour fidelity, which is a critical requirement in clinical workflows.
Through systematic ablation and data-starvation analyses, we show that foundation models confer benefits beyond absolute ac     curacy. They exhibit superior resilience in low-data regimes and reveal a non-trivial capacity–performance relationship characterized by early saturation and overfitting at extreme scales. These findings suggest that distilled, moderately sized transformer backbones provide the most favorable trade-off between accuracy, robustness, and computational efficiency for boundary-sensitive tasks like adnexal mass segmentation.

Taken together, our results provide preliminary empirical support for the integration of large-scale self-supervised vision models into clinical imaging workflows. By reducing reliance on extensive manual annotation while improving segmentation reliability, foundation-model-driven approaches represent a principled and scalable pathway toward robust, data-efficient medical image analysis in real-world clinical environments.